%% file: main.tex
\documentclass{article}

 \usepackage[preprint, nonanonymous]{neurips_2026}

\usepackage[utf8]{inputenc} 
\usepackage[T1]{fontenc}    
\usepackage[pagebackref,breaklinks,colorlinks,allcolors=nipsblue]{hyperref}
\usepackage{url}            
\usepackage{booktabs}       
\usepackage{amsfonts}       
\usepackage{nicefrac}       
\usepackage{microtype}      
\usepackage[table]{xcolor}

\usepackage{graphicx}
\usepackage{subfigure}
\usepackage{multirow}

\usepackage{amsmath}
\usepackage{array}
\usepackage{color}
\usepackage{bbding} 
\usepackage{algorithm}
\usepackage{algpseudocode}
\usepackage{stfloats}
\usepackage[none]{hyphenat}
\usepackage[most]{tcolorbox}
\usepackage{makecell}

\definecolor{nipsblue}{rgb}{0.35,0.49,0.74}
\definecolor{ggray}{RGB}{127,127,127}

\title{ChronoPhyBench: Do MLLMs Truly Understand the World or Merely Exploit Language Priors?}

\author{%
  Bin Zhu\textsuperscript{1,2,3$\ast$}, 
  Yanhao Jia\textsuperscript{4,$\ast$}, 
  Kexin Zhao\textsuperscript{1,$\ast$}, 
  Jie Wang\textsuperscript{2,5}, 
  Munan Ning\textsuperscript{1,2}, 
  Hao Li\textsuperscript{1,2}, 
  Yuwei Niu\textsuperscript{1}, 
  Tanqing Sun\textsuperscript{1}, \\
  \textbf{Huangchong Yan\textsuperscript{1}, 
  Mingjun Pan\textsuperscript{1}, 
  Xinyi Wu\textsuperscript{4}, 
  Qishen Yin\textsuperscript{1}, 
  Yunyang Ge\textsuperscript{1}, 
  Shuai Zhao\textsuperscript{4}, 
  Li Yuan\textsuperscript{1,2,$\dagger$}} \\ \\
  \textsuperscript{1}Peking University, Shenzhen Graduate School, 
  \textsuperscript{2}Peng Cheng Laboratory 
  \textsuperscript{3} Rabbitpre Intelligence
  \\
  \textsuperscript{4}Nanyang Technological University 
  \textsuperscript{5}Tsinghua University, 
   \\
  \texttt{\{binzhu\}@stu.pku.edu.cn}
  \texttt{\{yuanli-ece\}@pku.edu.cn} \\
  \textsuperscript{$\ast$}Equal contribution, \textsuperscript{$\dagger$}Corresponding author,  
}

\begin{document}

\maketitle

\input{sections/abstract}

\input{sections/introduction}
\input{sections/related_works}

\input{sections/method}
\input{sections/dataset}
\input{sections/experiment}
\input{sections/conclusion}

{
    \newpage
    \small
    \bibliographystyle{unsrt}
    \bibliography{reference}
}

{
    \newpage
    \appendix
    \input{sections/supp}
    \clearpage
}

\newpage
\input{checklist.tex}

\end{document}

%% file: sections/abstract.tex
\begin{abstract}
Recent advancements in Multimodal Large Language Models~(MLLMs) have demonstrated remarkable proficiency in open-world reasoning and understanding.
However, a critical ambiguity persists: it remains unclear whether these models genuinely synthesize cross-modal information to construct physically grounded reasoning chains, or if they merely exploit strong language priors to mask single-modality reliance, thereby hallucinating advanced multimodal capabilities.
Motivated by this, and to rigorously mitigate language modality bias and shortcuts, we propose a novel multimodal \textbf{Chrono}logical \textbf{Phy}sical Dynamics Reasoning Benchmark \textbf{ChronoPhyBench}, which unifies next state prediction with Visual Question Answering~(VQA) paradigms by conditioning on historical video context and textual captions to enforce models to deduce subsequent physical states through both single image selection and the inherently more complex task of multiple frame chronological sorting.
Concurrently, we construct a large-scale multimodal reasoning dataset curated using the ChronoPhyBench criteria, comprising over 10,000 long-form videos paired with meticulously annotated captions, totaling 5M tokens.
Our experimental evaluations reveal a stark contrast to conclusions drawn by previous benchmarks. The capacity of current open-source models to perform physically grounded multimodal reasoning remains in its infancy. Furthermore, we observe a concerning phenomenon where leading closed-source models frequently bypass visual reasoning entirely. They successfully predict correct answers relying solely on language priors, even when visual information is intentionally omitted or corrupted.
Ultimately, this work seeks to systematically stress-test the reasoning capabilities of multimodal models, quantify hallucination rates, and advance the development of Physical AI, thereby providing the community with a robust and transparent evaluation framework toward Artificial General Intelligence (AGI).
\end{abstract}

%% file: sections/introduction.tex
\section{Introduction}
\label{submission}
\vspace{-2mm}
Understanding physical processes is a fundamental capability for intelligent systems interacting with the real world. Predicting object motion, judging the physical plausibility of events, and reasoning about complex multi-object interactions all rely on the internal modeling of physical laws. Recent advances in Multimodal Foundation Models~(MLLMs) have demonstrated remarkable proficiency on physics-related tasks~\cite{black2025pi_, kim2024openvla, yuan2026fast, xing2025re}, with some models even achieving high success rates on open-world reasoning benchmarks~\cite{deng2025emerging, yu2025repa}. Nevertheless, a critical epistemic ambiguity remains: To what extent do these models truly synthesize cross-modal information to perform physically grounded reasoning, rather than merely exploiting strong language priors to "hallucinate" advanced capabilities?

Existing benchmarks typically rely on standard Visual Question Answering~(VQA) formats or frame-level success rates as their primary metrics~\cite{li2023weakly}. However, these paradigms struggle to distinguish whether the MLLMs are conducting genuine multimodal reasoning or simply use single modality prior as a shortcut~\cite{wang2026vfatbenchmarkingvisualfidelity, liang2024mitigating, qian2025decalign}. Because modern foundation models possess exceptionally strong textual priors, they often exploit statistical correlations in the text prompt or the question formulation to guess the correct answer~\cite{zhang2025robust, zhou2025mitigating, li2026quantum}. In doing so, they mask a profound reliance on a single modality (typically text) and exhibit a modality bias. For instance, the MLLMs might correctly predict the output of a physical collision not by analyzing the visual kinematics of the scene, but by matching frequent text patterns associated with the objects involved. This language-prior bias and vulnerability to modality conflict make it exceedingly difficult to pinpoint the true robustness and boundaries of current AI capabilities in physical intelligence~\cite{zhu2024unraveling, shu2025large, guo2025r}.

\begin{figure}[!t]
    \centering
    \includegraphics[width=1\linewidth]{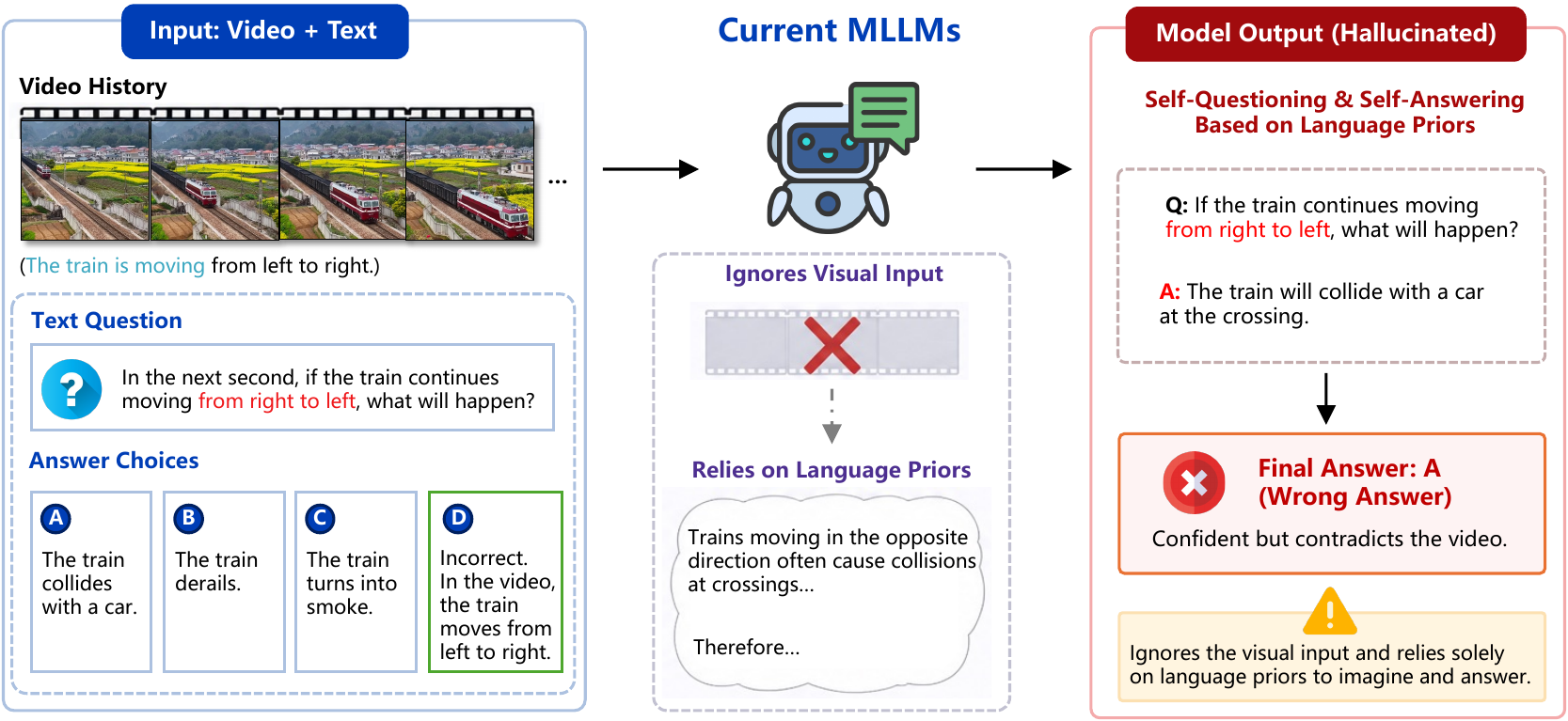}
    \vspace{-6mm}
    \caption{Illustration of hallucination in current Multimodal Foundation Models~(MLLMs). When presented with a text question that explicitly contradicts the visual video history (e.g., the text states that a train is moving from right to left, whereas the video shows it moving left to right), the model frequently ignores the visual input. Instead, it relies heavily on strong language priors to engage in self-questioning and self-answering, ultimately producing a confident yet entirely hallucinated prediction, such as predicting a collision rather than correctly identifying the textual error based on the visual evidence.}
    \label{fig:main}
    \vspace{-8mm}
\end{figure}

Initially, our evaluations of state-of-the-art MLLMs on standard physical Question Answering (QA) benchmarks yielded surprisingly high performance, often nearing human-level accuracy. However, this apparent success prompts a critical epistemic inquiry: do these models genuinely comprehend physical laws through visual grounding, or are they simply exploiting statistical language priors to "guess" the answers? To investigate this, we introduced systematic modality perturbations by crafting text prompts that explicitly conflict with the video context. The results were revealing; even top-tier models frequently bypass visual reasoning entirely, providing confident answers based solely on textual cues even when they directly contradict physical reality. This confirms that current MLLMs exhibit a profound lack of vision-centric physical understanding, relying instead on "modality laziness" to simulate reasoning capabilities.

To systematically expose this bottleneck and force models beyond their textual comfort zones, we introduce \textbf{ChronoPhyBench}. Departing from conventional VQA formulations that are easily bypassed by language shortcuts, ChronoPhyBench unifies next-state prediction with chronological physical reasoning through two rigorous paradigms. First, \textbf{Next-State Frame Selection} requires the model to identify the single physically consistent subsequent frame from a set of distractors that, while semantically similar, exhibit blatant physical violations such as incorrect trajectories or momentum. Second, \textbf{Multi-Frame Chronological Sorting} challenges the model to reorder a sequence of shuffled future states into their correct temporal and physical progression, a task that demands an internal modeling of global physical invariants over time. By increasing combinatorial complexity and requiring precise alignment with visual dynamics, these paradigms are explicitly designed to penalize reliance on language priors. This framework forces models to engage in genuine cross-modal synthesis, providing a robust baseline to determine if a model can truly perceive and reason about the physical laws governing the world.

In summary, this work provides a comprehensive operational definition and a robust evaluation baseline for physical intelligence under modality conflicts. Our main contributions are threefold:

\noindent 1. We identify and systematically expose the prevalence of shortcut learning and modality bias in MLLMs, demonstrating how textual priors induce hallucinations of physical reasoning capabilities.
    
\noindent 2. We introduce ChronoPhyBench, a unified benchmark that employs next-state prediction and chronological sorting to explicitly evaluate cross-modal synthesis and to penalize reliance on a single modality.
    
\noindent 3. We release ChronoPhy, a large-scale dataset of over 10,000 annotated videos, providing the community with a rigorous framework for stress-testing multimodal robustness and advancing the development of Physical AI toward genuine Artificial General Intelligence.

%% file: sections/related_works.tex
\vspace{-2mm}
\section{Related Work}
\vspace{-3mm}
\subsection{MLLM Benchmarks}
\vspace{-2mm}
In recent years, MLLMs have made significant strides in image and video understanding, catalyzing the development of diverse evaluation benchmarks. In the realm of image understanding, MME~\cite{fu2023mme} has established a large-scale evaluation system that systematically assesses model performance across perception and cognition task groups, including object existence, counting, and logical reasoning. MM-Vet~\cite{yu2024mm} focuses on evaluating the integration of multiple core capabilities—such as recognition, spatial awareness, logical reasoning, and mathematical calculation—within complex tasks. To challenge the depth of specialized knowledge, MMMU~\cite{yue2024mmmu} utilizes 11,500 questions spanning college-level expertise to test the advanced reasoning capabilities of multimodal models across interdisciplinary fields like science and engineering. Furthermore, MMBench~\cite{liu2024mmbench} provides a systematic diagnosis of capabilities such as object localization and social interaction through a fine-grained taxonomy of 20 capability dimensions. For short video understanding, datasets such as MMBench-Video~\cite{fang2024mmbench} and MVBench~\cite{li2024mvbench} provide large numbers of short-video question-answering tasks, assessing models’ abilities in semantic understanding, action recognition, and temporal reasoning. However, the videos in these datasets are typically under one minute to a few minutes long, which limits their ability to evaluate long-term video comprehension. For long video understanding, benchmarks such as LVBench~\cite{wang2025lvbench}, LongVideoBench~\cite{wu2024longvideobench}, Video-MME~\cite{fu2025video}, and MLVU~\cite{zhou2025mlvu} employ longer video segments to test models, challenging their long-term memory and multimodal reasoning capabilities. The average video length in these datasets often exceeds 10 minutes and can reach one hour or more. 
\vspace{-2mm}
\subsection{Embodied Intelligence Benchmarks}
\vspace{-2mm}
The advancement of Embodied Intelligence is fundamentally driven by the availability of comprehensive and challenging benchmarks. Early benchmarks, such as Habitat~\cite{liu2022behaviorhabitat20simulatorindependent} and ALFRED~\cite{ALFRED20}, primarily focused on basic navigation and simplistic object manipulation. Recently, however, the research focus has shifted toward complex semantic understanding and physical reasoning. For instance, A4Bench~\cite{wang2025affordance} systematically evaluates an agent's capability in affordance perception, emphasizing the transition from visual recognition to functional understanding. PhysToolBench~\cite{zhang2025phystoolbench} further addresses the gap in physical tool manipulation by challenging models across hierarchical levels, including recognition, functional analysis, and creative tool use, thereby testing the limits of their reasoning regarding physical commonsense, such as mechanics and material properties~\cite{qin2025robofactory, puyin2025quantiphy, motamed2026generative}. Moreover, comprehensive benchmarks like EmbodiedBench~\cite{yang2025embodiedbench} utilize multi-dimensional metrics to provide fine-grained evaluations across the entire pipeline, from high-level semantic planning to low-level atomic actions.
\vspace{-2mm}
\subsection{Physical Understanding Benchmarks}
\vspace{-2mm}


Physical reasoning and intuitive physics understanding are important directions in artificial intelligence research. To evaluate models’ capabilities in perceiving and reasoning about the physical world, numerous benchmark datasets have been proposed. Early benchmarks such as IntPhys~\cite{riochet2021intphys} and Clevrer~\cite{yi2019clevrer} primarily use simple geometric objects and simulated scenes, focusing on core physical principles such as object permanence, stability, collisions, and occlusion, emphasizing predictive reasoning abilities. Subsequent benchmarks, including Physion~\cite{bear2021physion}, Physion++~\cite{tung2023physion++}, ComPhy~\cite{chen2022comphy}, and ContPhy~\cite{zheng2024contphy}, increase the complexity of scenes, the number of objects, and types of physical interactions, supporting tasks such as trajectory prediction, physics-constrained reasoning, and continuous-time simulation. A4Bench~\cite{wang2025affordance} and PhysToolBench~\cite{zhang2025phystoolbench} further incorporate tool use and action sequences, requiring models to learn object affordances and handle physical causality closer to real-world scenarios. Question-answering-based benchmarks such as UGPhysics~\cite{xu2025ugphysics} and PHYBench~\cite{qiu2025phybench} are composed of university-level or competition-style physics problems, designed to evaluate models’ physical reasoning abilities. PhysBench~\cite{chow2025physbench} provides a wide range of physical scenarios and tasks, including collision prediction, stability judgment, object motion reasoning, and spatial relationship understanding, aiming to comprehensively assess models’ cognition and reasoning over the physical world.

%% file: sections/method.tex
\vspace{-3mm}
\section{Chronological Physical Dynamics Reasoning Benchmark}
\vspace{-3mm}
\begin{table}[!t]
    \centering
    \renewcommand{\arraystretch}{1.1}  
    \resizebox{\textwidth}{!}{%
        \begin{tabular}{@{} l c c c c c @{}}
            \toprule
            \makecell[l]{\textbf{Dataset /} \\ \textbf{Benchmark}} & 
            \makecell[c]{\textbf{Real-World} \\ \textbf{Dynamics}} & 
            \makecell[c]{\textbf{Next-State} \\ \textbf{Prediction}} & 
            \makecell[c]{\textbf{Chronological} \\ \textbf{Sorting}} & 
            \makecell[c]{\textbf{Hallucination} \\ \textbf{Decoys}} & 
            \makecell[c]{\textbf{Visual-Blind} \\ \textbf{Test}} \\
            \hline
            \rowcolor{gray!20}\multicolumn{6}{c}{\it{\textit{Physics-Centric (Synthetic)}}} \\
            \hline
            CATER~\cite{girdhar2019cater} & $\times$ & $\times$ & $\times$ & $\times$ & $\times$ \\
            CLEVRER~\cite{yi2019clevrer} & $\times$ & $\checkmark$ & $\times$ & $\times$ & $\times$ \\
            Physion~\cite{bear2021physion} & $\times$ & $\checkmark$ & $\times$ & $\times$ & $\times$ \\
            IntPhys 2~\cite{bordes2025intphys2benchmarkingintuitive} & $\times$ & \checkmark & $\times$ & $\times$ & $\times$ \\
            ComPhy~\cite{chen2025compositional} & $\times$ & \checkmark & $\times$ & $\times$ & $\times$ \\
            CRAFT~\cite{ates2022craft} & $\times$ & \checkmark & $\times$ & $\times$ & $\times$ \\
            \hline
            \rowcolor{gray!20}\multicolumn{6}{c}{\it{\textit{General Video MLLM (Real-world)}}} \\
            \hline
            Video-MME~\cite{fu2025video} & $\checkmark$ & $\times$ & $\times$ & $\times$ & $\times$ \\
            EgoSchema~\cite{mangalam2023egoschema} & $\checkmark$ & $\times$ & $\times$ & $\times$ & $\times$ \\
            STAR~\cite{wu2024star} & $\checkmark$ & $\checkmark$ & $\times$ & $\times$ & $\times$ \\
            Perception Test~\cite{patraucean2023perception} & $\checkmark$ & $\checkmark$ & $\times$ & $\times$ & $\times$ \\
            \midrule
            \textbf{ChronoPhy~(Ours)} & \textbf{\checkmark} & \textbf{\checkmark} & \textbf{\checkmark} & \textbf{\checkmark} & \textbf{\checkmark} \\
            \bottomrule
        \end{tabular}%
    }
    \caption{Feature comparison between our proposed benchmark and existing video/physical reasoning datasets. Our benchmark uniquely introduces modality stress-testing and temporal sorting to rigorously evaluate multimodal alignment.}
    \vspace{-8mm}
    \label{tab:dataset_feature_comparison}
\end{table}

Understanding the physical world requires models to synthesize complex visual dynamics rather than simply relying on static pattern recognition or textual priors. However, current multimodal evaluation paradigms often fail to distinguish between genuine physically grounded reasoning and superficial language-driven shortcut learning. To systematically quantify the true physical reasoning capabilities of MLLMs and expose their vulnerability to modality bias, we introduce ChronoPhyBench. Unlike traditional knowledge-based taxonomies, ChronoPhyBench is an operationally defined, task-driven framework. We abandon conventional static VQA formats in favor of predictive and chronological tasks that explicitly penalize reliance on single-modality priors.
\vspace{-2mm}
\subsection{Breaking Modality Shortcuts}
\vspace{-2mm}
The core vulnerability of current MLLMs lies in their tendency to bypass visual perception when strong language priors are present, a form of modality shortcut bias. If a physical question can be answered by merely associating textual keywords (e.g., matching "apple" + "drop" to "falling"), the visual input becomes redundant. To enforce cross-modal synthesis, ChronoPhyBench shifts the evaluation focus from describing the present to predicting the future and reconstructing the past. By conditioning tasks on extended video contexts, we force models to engage more deeply with visual kinematics, rendering text-only guessing statistically unprofitable.
\vspace{-2mm}
\subsection{Task I: Next-Frame Predictive Selection}
\label{subsec:taskI}
\vspace{-2mm}
This task evaluates the model's grasp of local causal dynamics. Given a historical video sequence depicting a physical event up to time $t$ and a textual description of the initial conditions, the model must select the correct immediate subsequent state at time $t+1$ from a set of visually plausible distractors. The distractors are carefully curated to share identical semantics but violate basic physical laws (e.g., incorrect trajectories, unbalanced forces, or impossible deformations). This paradigm ensures that models cannot infer the correct answer from text alone; they must extrapolate visual motion vectors to identify the physically sound outcome.
\vspace{-2mm}
\subsection{Task II: Chronological Multi-Frame Sorting}
\label{subsec:taskII}
\vspace{-2mm}
While Task I focuses on instantaneous causality, Task II evaluates long-term global physical consistency. The model is presented with the initial video context and a set of randomly shuffled image frames representing distinct, non-consecutive future states of the physical system. The objective is to reconstruct the correct chronological sequence of these frames. This represents a significantly higher level of physical reasoning, requiring the model to internalize latent invariant variables (e.g., conservation of momentum or energy) and to map a continuous causal chain across multiple spatial-temporal windows, thereby severely penalizing random guessing and hallucinations.
\vspace{-2mm}
\subsection{The Modality Stress-Testing Framework}
\label{subsec:stress-test}
\vspace{-2mm}
The most critical feature of ChronoPhyBench is its built-in mechanism for quantifying hallucination rates and reliance on a single modality. Inspired by the framework of \cite{jia2025seeingsoundhearingsight, ling2025vmbench, zhou2025pai, wang2025physunibench}, we introduce a systematic stress-testing protocol comprising three distinct evaluation settings:

\noindent \textbf{Standard Multimodal Setting}: Models receive both the full video context and the descriptive text prompt, serving as the control group to evaluate the model's fundamental capability.

\noindent \textbf{Visual-Blind Setting (Textual Prior Test)}: The visual input is intentionally completely omitted, forcing the model to answer based solely on the text. High performance in this setting explicitly exposes severe dataset bias and the model's reliance on text shortcuts.

\noindent \textbf{Conflict Setting}: The visual context is retained, while the textual prompt is altered to describe a physically incorrect outcome. This tests the model's robustness and whether it relies on its visual perception rather than deceptive text priors.

By comparing performance across these three settings, ChronoPhyBench provides the first quantitative metric for multimodal reasoning bias, clearly delineating models that genuinely understand physics from those that merely simulate it via text.

%% file: sections/dataset.tex
\begin{figure}[!t]
    \centering
    \includegraphics[width=\linewidth]{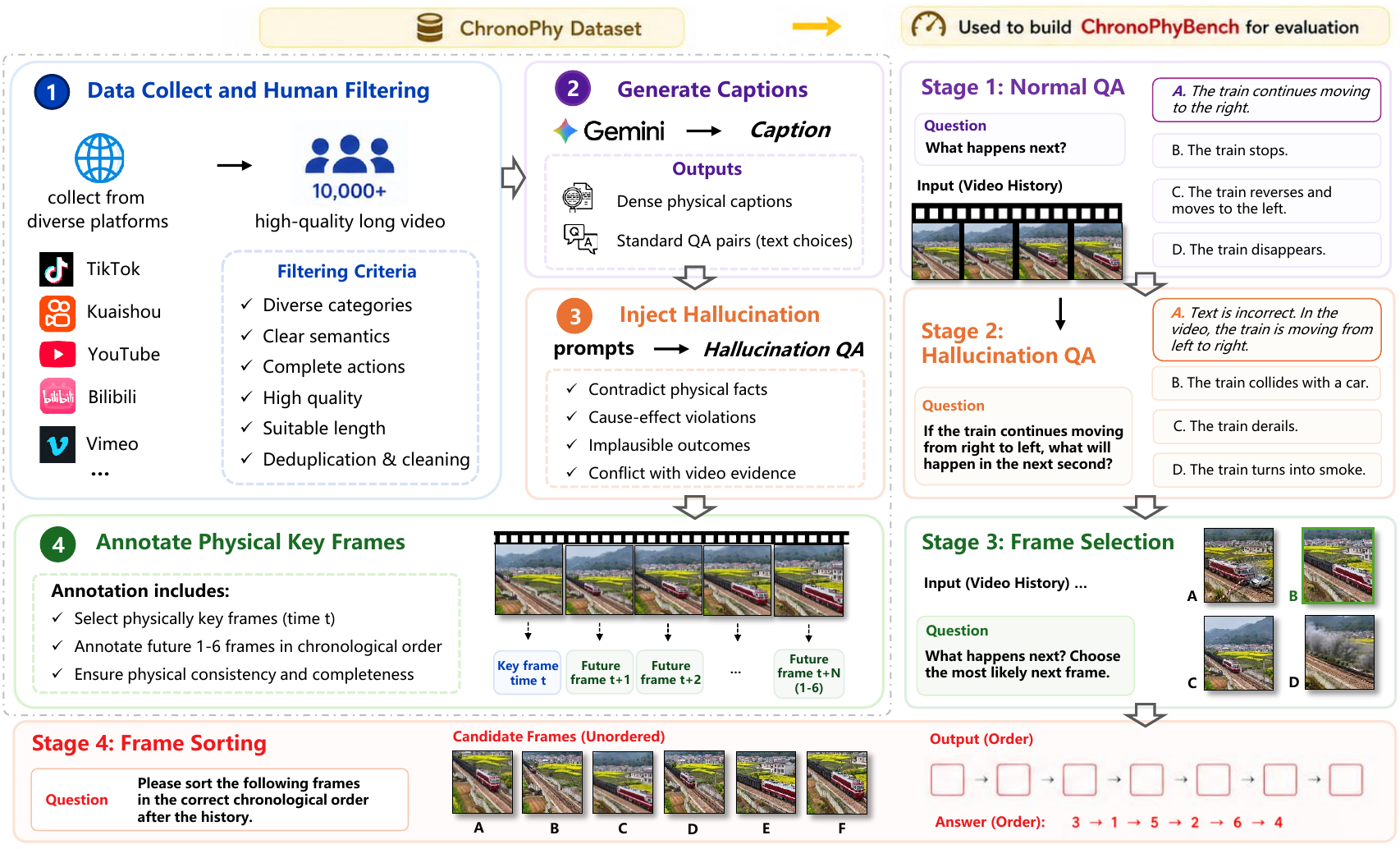}
    \vspace{-6mm}
    \caption{The overall construction pipeline of the ChronoPhy dataset and benchmark. The process begins with the collection and rigorous human filtering of over 10,000 high-quality, real-world videos from diverse platforms. Next, Gemini is utilized to generate dense physical captions and standard Normal QA pairs (Stage 1). To explicitly stress-test model robustness, adversarial prompts are systematically injected to create Hallucination QA pairs (Stage 2). Finally, human annotators carefully select physical keyframes to construct the core visually grounded evaluation paradigms: Next-State Frame Selection (Stage 3) and Multi-Frame Chronological Sorting (Stage 4).}
    \label{fig:dataset}
    \vspace{-5mm}
\end{figure}
\vspace{-2mm}
\section{The ChronoPhy Dataset: Construction and Statistics}
\vspace{-3mm}
To support the rigorous evaluation protocols of ChronoPhyBench, we introduce \textbf{ChronoPhy}, a large-scale, high-fidelity multimodal dataset. The construction pipeline integrates automated MLLM generation with rigorous human-in-the-loop curation to ensure physical accuracy and to intentionally stress-test modalities~\cite{achiam2023gpt, team2023gemini}. The final ChronoPhy dataset represents a comprehensive and challenging testbed for physical AI. The precise statistical distribution is detailed in Tab.\ref{tab:dataset_stats}.

\begin{table}[!t]
  \centering
  \renewcommand{\arraystretch}{1.1}  
  \begin{tabular}{@{} l r p{8cm} @{}}
    \toprule
    \textbf{Data Category} & \textbf{Count} & \textbf{Description / Purpose} \\
    \midrule
    Total Curated Videos & 10,000+ & High-quality physical interaction videos forming the foundation. \\
    \midrule
    Standard QA Pairs & 10,000+ & Baseline multimodal comprehension (Aligned visual-text scenarios). \\
    Hallucination QA Pairs & 5,000+ & Stress-testing instances with injected modality conflicts. \\
    Chronological Sorting Pairs & 400+ & Multi-frame (2-6 frames) temporal and physical ordering sequences. \\
    Next-State Selection Pairs & 400+ & Single-frame predictive choices from $N$ distractors. \\
    \midrule
    \textbf{Total Predictive Pairs} & \textbf{16,000+} & \textbf{Combined subset dedicated to Tasks I and II.} \\
    \bottomrule
  \end{tabular}
  \caption{Statistics of the ChronoPhy Dataset}
  \label{tab:dataset_stats}
  \vspace{-2mm}
\end{table}

\vspace{-2mm}
\subsection{Data Collection and Human Annotation}
\vspace{-2mm}
The foundation of ChronoPhy comprises a diverse corpus of real-world video footage curated from open web repositories. A critical limitation of existing physical reasoning benchmarks is their over-reliance on idealized, physics-engine renderings (e.g., MuJoCo or Blender), as shown in Tab.~\ref{tab:dataset_feature_comparison}. In contrast, ChronoPhy champions real-world physical complexity. Our videos naturally capture intricate, high-fidelity dynamics, such as non-rigid deformations, fluid viscosity, and unpredictable environmental resistance, which synthetic environments struggle to simulate accurately. To ensure data quality and causal clarity, human experts applied a rigorous filtering protocol, systematically discarding instances exhibiting severe motion blur, artificial editing, or ambiguous physical causality. The resulting corpus spans a comprehensive spectrum of physical phenomena, ranging from rigid-body collisions to complex mechanical linkages, thereby establishing an unprecedented, empirically grounded visual baseline for the benchmark.
\vspace{-3mm}
\subsection{Automated QA Generation and Human-Centric Refinement}
\vspace{-2mm}
Following video curation, we employed a scalable pipeline combining VLM-based generation with expert manual intervention to construct high-quality text-video pairs, as illustrated in Fig.~\ref{fig:dataset} (2):

\noindent \textbf{Stage I: Standard QA Construction and Verification.} For each video, the Gemini model was initially prompted to generate dense physical captions and corresponding visual-text Question-Answering (QA) pairs. To guarantee physical fidelity and linguistic precision, human experts with physics backgrounds conducted a rigorous multi-round review. This process involved pruning approximately 40\% of the initial candidates due to quality issues or factual inaccuracies. For the remaining samples, annotators meticulously corrected technical terminology and re-aligned texts that diverged from the visual ground-truth. These high-fidelity pairs establish a robust baseline for evaluating foundational multimodal comprehension.

\noindent \textbf{Stage II: Hallucination Injection and Expert Tuning.} To operationalize the \textit{Conflict Setting} (Sec.~\ref{subsec:stress-test}), we utilized a secondary prompting strategy where Gemini injected physically plausible but erroneous hallucinations into the standard QA pairs. Critically, these "trap" questions underwent a second round of human refinement. Nearly 45\% of the candidate hallucination pairs were discarded due to excessive ambiguity or lack of visual decidability. Experts manually adjusted the remaining samples to ensure they were sufficiently deceptive yet logically identifiable through visual evidence, creating a robust "Hallucination QA" subset that effectively exposes single-modality bias.

\vspace{-2mm}



\vspace{-2mm}
\subsection{Keyframe Curation for Predictive and Sorting Tasks}
\vspace{-2mm}

To support the core predictive reasoning paradigms of ChronoPhyBench  (Tasks I and II), human annotators manually reviewed a subset of complex videos to extract "physical keyframes." These keyframes capture critical moments of state transition (e.g., the exact moment of impact, the apex of a trajectory). Using an initial keyframe as the contextual anchor, we constructed two distinct task sets:

\textbf{Multi-Frame Next-State Selection~(Frame Selection)}: Annotators defined an initial state and models are required to select the immediate subsequent physical state from $N$ visually plausible distractor frames.

\textbf{Multi-Frame Chronological Sorting~(Frame Sorting)}: Annotators extracted a sequence of 2 to 6 non-consecutive future frames from a pool of $N$ possible options. Models must arrange these frames to reconstruct the event's correct temporal and physical evolution.

%% file: sections/experiment.tex
\section{Experiments}
\vspace{-3mm}
To comprehensively evaluate physically grounded reasoning and modality robustness, we benchmark 25 diverse MLLMs, comprising 14 open-source models, 3 reasoning-focused models, and 8 state-of-the-art closed-source models. For our evaluations, we sample representative subsets from ChronoPhyBench: 10,000 instances for Standard QA, 5,000 for Hallucination QA, and 400 each for the Next-State Selection and Chronological Sorting tasks. To ensure strict fairness and statistical reliability, all models are evaluated three times independently in a zero-shot setting, and the average performance is reported. Furthermore, we use a standardized prompt template and a unified set of generation hyperparameters across the entire model suite, ensuring that performance discrepancies reflect inherent multimodal alignment capabilities rather than hyperparameter-tuning advantages. For more details, please ref to \textbf{Supp Sec.~\ref{supp_sec:setting}}.
\vspace{-2mm}
\subsection{Analysis on Frame Selection}
\vspace{-2mm}
\input{sections/table}

\textbf{Performance decreases as the number of frame choices increases}.
The most striking observation across the benchmark is the severe performance degradation when models transition from the 3-choice to the 6-choice setting. This phenomenon confirms that many models rely on superficial language priors rather than robust physical understanding. For instance, InternVL-3 38B drops precipitously from 39\% to 16.2\%, representing a collapse of nearly 23 percentage points and placing it entirely at the random-chance baseline (16.67\%) for the 6-choice task. Similarly, Kimi-VL A3B-Instruct drops from 36.82\% to 20.74\%. This indicates that when the combinatorial space of visual distractors expands, models that rely on "shortcut learning" are easily deceived by physically impossible yet visually similar decoy frames.

\textbf{Closed-source models outperform open-source models}.
A clear stratification exists within the model hierarchy. Leading proprietary models demonstrate a profound resilience to physical hallucination. GPT-5.5 establishes a dominant ceiling, achieving 72.5\% on Acc@3 and remarkably maintaining 70.5\%  on Acc@6, proving immunity to the increased distractor count. Seed-2.0 Pro-260215 (68.34\% / 65.58\%) and Gemini-3.0 Pro-Preview (68.87\% / 63.13\%) also exhibit robust cross-modal physical synthesis.  Conversely, the open-source community faces a steep "reasoning-to-agency" watershed. Nearly all evaluated open-source models, including InternVL-3.5 14B (14.97\%) and Qwen3.5-9B (14.49\%), plummet below the 16.67\% random guessing threshold on the 6-choice task. Furthermore, InternVL-3.5 38B achieves only 30.12\%  on the 3-choice task, failing to surpass the 33.3\% random baseline, suggesting that these models are actively misled by conflicting multimodal inputs rather than just guessing blindly.

\textbf{Benefits of CoT}.
An interesting secondary finding is the relative robustness of models equipped with Chain-of-Thought (CoT) or "thinking" mechanisms. Despite having fewer parameters, GLM-4.1V 9B-Thinking achieves 50.19\% in Acc@3 and maintains 39.35\% in Acc@6, outperforming the legacy proprietary model GPT-5.3 Chat-Latest, which scores 45.95\% and 34.26\%, respectively. This suggests that explicitly forcing models to allocate intermediate computational steps can partially mitigate modality bias and improve grounding in physical causality.
\vspace{-2mm}
\subsection{Analysis on Frame Sorting}
\vspace{-2mm}
\textbf{Multi-Frame Sorting still challenges all MLLMs}.
The transition from 3-frame sorting (6 possible permutations) to 4-frame sorting (24 possible permutations) triggers a severe "combinatorial collapse" in performance across all models. Even the state-of-the-art closed-source models suffer catastrophic degradation. For example, Gemini-3.0 Pro-Preview drops from 70\% to 35.5\%, a staggering absolute decrease of 34.5\%. Similarly, GPT-5.5 plummets from 72\% to 40.5\%. This suggests that while top-tier models can deduce immediate local causality, their ability to maintain a coherent, global physical invariant across an extended temporal window remains highly brittle.

\textbf{Open-Source Models Regress to Random Guessing}.
The disparity between closed-source and open-source models is particularly pronounced in the 4-frame sorting task. The mathematical random baseline for ordering 4 frames is roughly 4.17\%. A significant portion of the open-source community effectively regresses to this baseline. Models such as InternVL-3 8B, InternVL-3.5 8B, and MiniCPM-V-4.5 all bottom out at exactly 5\% accuracy in the 4-frame sorting task. Even larger parameter models, such as InternVL-3.5 38B, achieve only 5.5\%. This indicates a complete failure of long-horizon kinematic modeling, demonstrating that these models rely on transient visual cues rather than internalizing continuous physical dynamics.

\textbf{Those SOTA models also have Anomalies}.
Interestingly, while Seed-2.0 Pro-260215 achieves the highest accuracy in the 3-frame sorting task at 74\%, its predecessor, Seed 1.8-251228, demonstrates remarkable stability. The latter achieves 61\% accuracy on the 3-frame task and retains 46.5\% on the 4-frame task, securing the highest absolute performance in the more complex evaluation. This anomaly highlights that scaling alone does not guarantee proportional robustness to temporal modality conflicts; specific architectural alignments or training data distributions that involve video interpolation and physical dynamics play a more critical role.

For more experiment results and analysis, please refer to \textbf{Supp Sec.~\ref{suppsec:exp}} and \textbf{Tab.~\ref{tab:qa_results}}.

\vspace{-2mm}
\subsection{Visualization Result}
\vspace{-2mm}

\begin{figure}
    \centering
    \includegraphics[width=\linewidth]{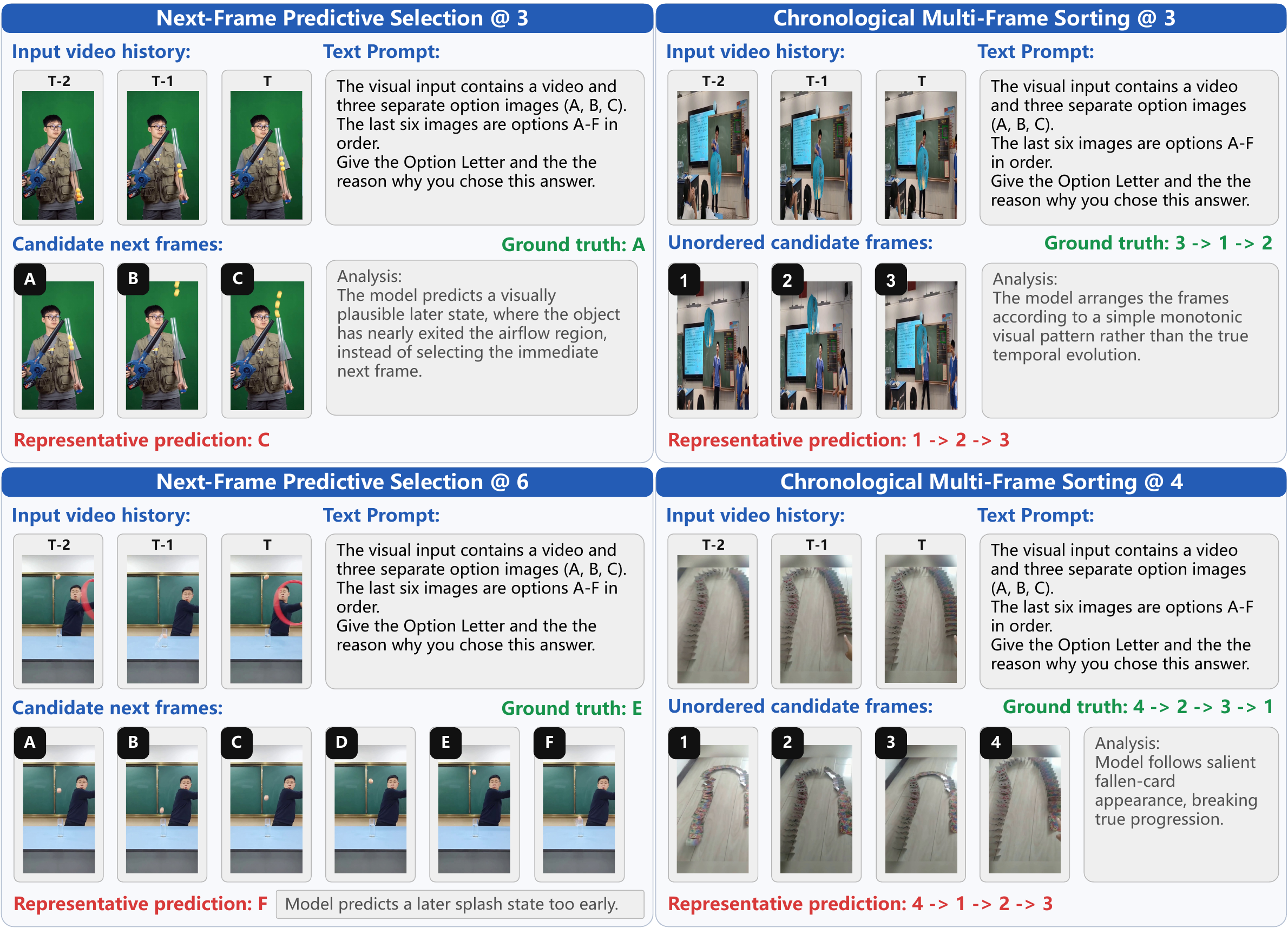}
    \vspace{-6mm}
    \caption{Some visualization results from ChronoPhyBench. The left column illustrates Next-Frame Predictive Selection, while the right column demonstrates Chronological Multi-Frame Sorting.}
    \label{fig:vis}
    \vspace{-5mm}
\end{figure}

To better understand the specific failure modes of current MLLMs, we conduct a qualitative analysis of their predictions on ChronoPhyBench (Fig.~\ref{fig:vis}). The visualizations reveal two prominent, systematic errors that quantitatively lower model performance: 

\textbf{The Temporal Skip in Predictive Selection}. In the Next-Frame Predictive Selection tasks (Left Part of Fig.~\ref{fig:vis}), models exhibit a pervasive inability to maintain fine-grained temporal resolution. Rather than identifying the immediate, causally adjacent next frame ($t+1$), models frequently jump ahead to select a visually salient, but temporally distant, future state ($t+k$). For instance, in the 3-choice hairdryer scenario, the model ignores the immediate onset of paper fluttering and instead selects a frame in which the paper has already been blown far away. Similarly, in the 6-choice water droplet task, the model prematurely predicts a fully formed splash (Frame F) rather than the subtle initial surface impact (Frame E). This indicates that while MLLMs can associate semantic concepts (e.g., "hairdryer causes object to move," "drop causes splash"), they fundamentally lack the kinematic grounding required to predict immediate, step-by-step physical evolution. They recognize the final outcome but fail to trace the local causal chain.  

\textbf{The Monotonic Fallacy in Chronological Sorting}. The Multi-Frame Sorting tasks (Right Part of Fig.~\ref{fig:vis}) expose an even more severe limitation in global physical reasoning. When tasked with reordering a scrambled sequence of frames, models often default to sorting by superficial, monotonic visual changes—such as the amount of area moved or the apparent size of an object—rather than by true physical progression. In the domino example, the model correctly identifies the first fallen state but then completely breaks the true causal progression, seemingly confused by the salient appearance of the fallen cards and failing to track the continuous transfer of kinetic energy. This "monotonic fallacy" demonstrates that MLLMs primarily engage in simplistic visual pattern matching rather than internalizing the complex underlying physical invariants (such as momentum and energy conservation) required to reconstruct a coherent chronological timeline.

%% file: sections/table.tex
\definecolor{navy_blue}{RGB}{0, 47, 167}
\definecolor{richCrimson}{RGB}{153, 0, 51}
\definecolor{scoreGreen}{RGB}{0, 128, 0}

\newcommand{\gbf}[1]{\textcolor{scoreGreen}{\textbf{#1}}}
\newcommand{\rbf}[1]{\textcolor{richCrimson}{\textbf{#1}}}

\begin{table}[!t]
    \centering
    \renewcommand{\arraystretch}{1.1}
    \setlength{\tabcolsep}{3.5mm}
    \begin{tabular}{@{} c | c >{\centering\arraybackslash}p{2.5cm} | c >{\centering\arraybackslash}p{2.2cm} @{}}
        \toprule
        \multirow{2}{*}{\textbf{Model}} & \multicolumn{2}{c|}{Frame Selection} & \multicolumn{2}{c}{Frame Sorting} \\
        \cmidrule(rl){2-3}\cmidrule(rl){4-5}
        & \textbf{Acc@3} & \textbf{Acc@6} & \textbf{Kframe@3} & \textbf{Kframe@4} \\
        \noalign{\hrule height 1.5pt}
        \textit{Random Baseline} & 33.33 & 16.67$_{\gbf{(-16.66)}}$ & 16.67 & 4.17$_{\gbf{(-12.5)}}$ \\
        \hline
        \rowcolor{gray!20}\multicolumn{5}{c}{\it{\textbf{Open-Source Models}}} \\
        \hline
        Qwen3.5 9B & - & 14.49 & - & - \\
        InternVL-3 38B & 41.0 & 16.0$_{\gbf{(-25.0)}}$ & 23.5 & 8.0$_{\gbf{(-15.5)}}$ \\
        InternVL-3 14B & 40.5 & 18.5$_{\gbf{(-22.0)}}$ & 23.5 & 8.0$_{\gbf{(-15.5)}}$ \\
        Kimi-VL A3B-Instruct & 35.5 & 20.5$_{\gbf{(-15.0)}}$ & - & - \\
        InternVL-3.5 14B & 29.5 & 11.5$_{\gbf{(-18.0)}}$ & 22.5 & 8.0$_{\gbf{(-14.5)}}$ \\
        MiniCPM-V-4.5 & 33.5 & 18.0$_{\gbf{(-15.5)}}$ & 13.5 & 5.0$_{\gbf{(-8.5)}}$ \\
        MiMo-VL 7B-RL-2508 & 38.0 & 20.0$_{\gbf{(-18.0)}}$ & 39.0 & 15.0$_{\gbf{(-24.0)}}$ \\
        InternVL-3 8B & 37.5 & 23.5$_{\gbf{(-14.0)}}$ & 30.5 & 5.0$_{\gbf{(-25.5)}}$ \\
        Qwen3-VL 8B-Instruct & 38.5 & 22.5$_{\gbf{(-16.0)}}$ & 31.5 & 14.0$_{\gbf{(-17.5)}}$ \\
        Ovis-2.5 9B & 41.0 & 28.5$_{\gbf{(-13.5)}}$ & 22.5 & 7.5$_{\gbf{(-15.0)}}$ \\
        InternVL-3.5 38B & 29.5 & 17.0$_{\gbf{(-12.5)}}$ & 28.0 & 5.5$_{\gbf{(-22.5)}}$ \\
        InternVL-3.5 8B & 30.0 & 19.5$_{\gbf{(-10.5)}}$ & 20.0 & 5.0$_{\gbf{(-15.0)}}$ \\
        Qwen3-VL 32B-Instruct & 37.0 & 25.5$_{\gbf{(-11.5)}}$ & 41.0 & 23.0$_{\gbf{(-18.0)}}$ \\
        GLM 4.6V-Flash & 50.0 & 45.0$_{\gbf{(-5.0)}}$ & 40.0 & 15.0$_{\gbf{(-25.0)}}$ \\
        \hline
        \rowcolor{gray!20}\multicolumn{5}{c}{\it{\textbf{Reasoning/Thinking Models}}} \\
        \hline
        GLM-4.1V-9B-Thinking & 52.5 & 39.5$_{\gbf{(-13.0)}}$ & 29.5 & 13.5$_{\gbf{(-16.0)}}$ \\
        Qwen3-VL-8B-Thinking & 40.0 & 25.0$_{\gbf{(-15.0)}}$ & 23.5 & 16.0$_{\gbf{(-7.5)}}$ \\
        Seed-1.6-Thinking-250715 & 41.5 & 38.5$_{\gbf{(-3.0)}}$ & 42.0 & 27.0$_{\gbf{(-15.0)}}$ \\
        \hline
        \rowcolor{gray!20}\multicolumn{5}{c}{\it{\textbf{Closed-Source Model}}} \\
        \hline
        Gemini-2.5 Flash & - & 37.85 & - & - \\
        Gemini-2.5 Pro & - & 46.63 & - & - \\
        GPT-5.4 & 52.5 & 37.5$_{\gbf{(-15.0)}}$ & 32.5 & 16.0$_{\gbf{(-16.5)}}$ \\
        GPT-5.3 Chat-Latest & 45.5 & 34.5$_{\gbf{(-11.0)}}$ & 53.0 & 33.5$_{\gbf{(-19.5)}}$ \\
        Gemini-3.0 Pro Preview & \underline{68.5} & 63.0$_{\gbf{(-5.5)}}$ & 70.0 & 35.5$_{\gbf{(-34.5)}}$ \\
        Seed-2.0 Pro-260215 & 68.0 & \underline{65.0}$_{\gbf{(-3.0)}}$ & \textbf{74.0} & \underline{45.5}$_{\gbf{(-28.5)}}$ \\
        Seed 1.8-251228 & 55.0 & 53.0$_{\gbf{(-2.0)}}$ & 61.0 & \textbf{46.5}$_{\gbf{(-14.5)}}$ \\
        GPT-5.5 & \textbf{72.5} & \textbf{70.5}$_{\gbf{(-2.0)}}$ & \underline{72.0} & 40.5$_{\gbf{(-31.5)}}$ \\
        \bottomrule
    \end{tabular}
    \caption{Performance comparison of MFMs on the ChronoPhyBench. Acc@3 and Acc@6 represent accuracy on the 3-selection and 6-selection paradigms, respectively. KFrame@3 and KFrame@4 represent accuracy on the 3-sorting and 4-sorting paradigms, respectively. In two tasks, the best is in bold and the second best is underlined.}
    \label{tab:predictive_results}
    \vspace{-4mm}
\end{table}

%% file: sections/conclusion.tex
\vspace{-5mm}
\section{Conclusion}
\vspace{-4mm}
In this work, we address the critical issue of the modality laziness in MLLMs by introducing ChronoPhyBench, a novel stress-testing framework that replaces static VQA with Next-State Predictive Selection, Multi-Frame Chronological Sorting, and adversarial hallucination decoys to enforce genuine visual grounding. Our extensive evaluations expose a profound disconnect between visual perception and physical causality in current state-of-the-art models: while leading proprietary models demonstrate baseline local reasoning, they suffer a severe "combinatorial collapse" in long-horizon sorting tasks, and the majority of open-source models regress entirely to random guessing when faced with dense visual distractors and temporal shifts. By systematically exposing how MLLMs bypass visual evidence in favor of deceptive language priors, ChronoPhyBench provides a rigorous crucible for the community to advance architectural innovations in temporal sequence modeling and explicit cross-modal alignment, ultimately serving as a crucial stepping stone toward physically grounded Artificial General Intelligence.

%% file: sections/supp.tex
\section{Experiment Settings}
\label{supp_sec:setting}
The experimental settings of this study are as follows: all test videos are several seconds in duration, with a uniform sampling strategy of 8 frames per video. Regarding model generation parameters, the Temperature for closed-source models is set to 0.1. For the open-source Qwen series, the Temperature and Top-p are set to 0.7 and 0.8, respectively, while for the GLM series, they are set to 0.8 and 0.6. As for the generation length, the Max Tokens is set to 8192 for reasoning (thinking) models and 2048 for non-reasoning models.
\section{Prompt Settings}

Frame selection prompt:

\begin{tcolorbox}[
    colback=gray!10,
    colframe=black,
    boxrule=0.6pt,
    arc=3pt,
    left=6pt,
    right=6pt,
    top=8pt,
    bottom=8pt,
    width=\linewidth,
    breakable
]
    \begin{quote}\small
        \textit{
            Note: The visual input contains a video and three separate option images (A, B, C). The last three images are options A, B, and C in order.
            I have provided them separately. Please do not confuse the option images with the video frames. \\
        }
        \par\smallskip
        \hspace*{1em}%
        \begin{minipage}{0.9\linewidth}
            \ttfamily
            \{ \\
            \hspace*{1.5em}"Question Context": {item['question']} \\
            \hspace*{1.5em}"Conclusion": Give the Option Letter. \\
            \hspace*{1.5em}"Analysis": Give the reason why you chose this answer. \\
            \}
        \end{minipage}
        
        \textit{
            You must follow the output format strictly. DO NOT provide extra explanations outside the format. \\
        }
    \end{quote}
\end{tcolorbox}

QA selection prompt:

\begin{tcolorbox}[
    colback=gray!10,
    colframe=black,
    boxrule=0.6pt,
    arc=3pt,
    left=6pt,
    right=6pt,
    top=8pt,
    bottom=8pt,
    width=\linewidth,
    breakable
]
    \begin{quote}\small
        \textit{
            Please analyze the physical process in this video and answer the following question: \\
        }
        \par\smallskip
        \hspace*{1em}%
        \begin{minipage}{0.9\linewidth}
            \ttfamily
            \{ \\
            \hspace*{1.5em}"Question Context": {item['question']} \\
            \hspace*{1.5em}"Options": {options\_str} \\
            \}
        \end{minipage}
        
        \textit{
            Please structure your response as follows:
        }
        \par\smallskip
        \hspace*{1em}%
        \begin{minipage}{\linewidth}
            \ttfamily
            \{ \\
            \hspace*{1.5em}"Conclusion": State the final answer by providing the letter AND the content of the chosen option (e.g., 'A. The ball will fall'). \\
            \hspace*{1.5em}"Analysis": Explain the physical principles (e.g., gravity, collision, conservation) that lead to the result. \\
            \}
        \end{minipage}
    \end{quote}
\end{tcolorbox}

\section{More Results}
\label{suppsec:exp}
As demonstrated in Table \ref{tab:qa_results}, evaluating models solely on Standard QA yields a misleadingly optimistic view of their physical reasoning capabilities. Without adversarial interventions, models like Qwen3-VL-32B-Instruct and GPT-5.4 achieve near-perfect scores of 97.35\% and 98.22\%, respectively. However, the introduction of modality conflicts via Hallucination QA exposes a severe reliance on textual priors across most architectures. Open-source models suffer catastrophic performance degradation; notably, GLM-4.6V-Flash and Qwen3-VL-8B-Instruct experience absolute drops ($\Delta$) of 46.11\% and 44.96\%, respectively. Even explicitly reasoning-focused models, such as GLM-4.1V-9B-Thinking, are not immune, dropping by 40.00\%. This stark contrast quantitatively confirms that these models frequently bypass visual evidence, defaulting to text-based statistical guessing when confronted with adversarial decoys. Conversely, the closed-source GPT-5.4 demonstrates remarkable multimodal robustness, maintaining 93.66\% accuracy with only a marginal degradation of 4.56\%, establishing a clear watershed between superficial pattern matching and genuinely grounded visual comprehension.

\begin{table}[ht]
    \centering
    \renewcommand{\arraystretch}{1.2}
    \setlength{\tabcolsep}{3mm}
    \begin{tabular}{@{} l | c c | c @{}}
        \toprule
        \textbf{Model} & \textbf{Standard QA (\%)} & \textbf{Hallucination QA (\%)} & \textbf{Visual-Blind (\%)} \\
        \hline
        \rowcolor{gray!20}\multicolumn{4}{c}{\it{\textbf{Open-Source Models}}} \\
        \hline
        Qwen3-VL-32B-Instruct & 97.35 & 69.35$_{\gbf{(-28.00)}}$ & 93.85$_{\gbf{(-3.50)}}$\\
        Qwen3-VL-8B-Instruct & 93.58 & 48.62$_{\gbf{(-44.96)}}$ & 90.01$_{\gbf{(-3.57)}}$\\
        GLM-4.6V-Flash & 90.11 & 44.00$_{\gbf{(-46.11)}}$ & 85.16$_{\gbf{(-4.95)}}$\\
        \hline
        \rowcolor{gray!20}\multicolumn{4}{c}{\it{\textbf{Reasoning/Thinking Models}}} \\
        \hline
        GLM-4.1V-9B-Thinking & 84.67 & 44.67$_{\gbf{(-40.00)}}$ & 78.61$_{\gbf{(-6.06)}}$\\
        \hline
        \rowcolor{gray!20}\multicolumn{4}{c}{\it{\textbf{Closed-Source Model}}} \\
        \hline
        GPT-5.4 & 98.22 & 93.66$_{\gbf{(-4.56)}}$ & 95.01$_{\gbf{(-3.21)}}$\\
        \bottomrule
    \end{tabular}
    \vspace{2mm}
    \caption{Performance comparison on Standard QA versus Hallucination QA and Visual-Blind condition. \gbf{X} represents the absolute performance drop when adversarial textual decoys (modality conflicts) are introduced.}
    \label{tab:qa_results}
\end{table}

\section{Declaration for ChronoPhy Dataset}
ChronoPhy is constructed from publicly accessible online videos and is intended solely for academic research on physically grounded multimodal evaluation. We filter out videos containing sensitive scenes, clearly identifiable private individuals, visible license plates, children, or other personally identifiable information whenever possible. When copyright or platform licensing conditions are unclear, we release annotations, metadata, and source references rather than redistributing raw videos, and we provide a takedown mechanism for content owners. Human annotators are provided with detailed guidelines and compensated according to local labor standards. Since part of the QA data is generated by a multimodal foundation model, ChronoPhy may inherit linguistic artifacts or generator-specific biases; we mitigate this through human verification and filtering. ChronoPhyBench is intended as a diagnostic benchmark and should not be interpreted as certifying real-world robotic safety or general physical intelligence.

\section{Broader Impacts, Limitations, and Future Work}

\subsection{Broader Social Impacts}
The deployment of MLLMs in embodied environments—such as autonomous driving and robotics—demands absolute reliability. By exposing the vulnerability of current models to text-induced hallucinations, ChronoPhyBench serves a critical safety function. Modality bias is not a fixed attribute; it varies dynamically with input feature integrity and environmental reliability. Uncovering these failure modes prevents the premature deployment of physical AI systems that might confidently hallucinate impossible actions when text priors are strong or visual signals are compromised.

\subsection{Limitations}
Despite the rigorous design of our stress-testing framework, this study presents several limitations:
\begin{itemize}

    \item \textbf{Generator Bias}: The initial generation of dense physical captions and Hallucination QA pairs relies heavily on a single proprietary model. This automated pipeline may inadvertently tailor the hallucination decoys to the specific failure modes of that generator, potentially skewing the benchmark's distribution and slightly biasing the evaluation against certain architectures.
    \item \textbf{Empirical Gaps in Ablation}: While the "Visual-Blind Setting" is conceptually introduced as a core mechanism to expose dataset bias and quantify modality laziness, the main empirical tables currently lack the full quantitative results for this specific setting, limiting the immediate validation of our modality bias claims.
    \item \textbf{Dataset Utilization}: Although the foundational dataset contains over 10,000 curated videos, the highly rigorous Frame Selection and Sorting evaluations utilize a highly distilled subset of approximately 1,000 tasks. The broader utility and public-release structure for the remaining thousands of standard QA pairs require further elaboration. 
    \item \textbf{Literature Contextualization}: The current discussion could be more thoroughly contextualized against the rapidly expanding body of literature on long-form video understanding, complex temporal reasoning, and multi-frame sequential evaluation.

\end{itemize}

\subsection{Future Work}
Subsequent iterations of this research will prioritize expanding the empirical suite to fully document the Visual-Blind evaluations across all model scales. To mitigate data-generation bias, we will transition to an ensemble-based MLLM generation pipeline to craft hallucination decoys. Furthermore, we plan to redefine modality alignment—moving away from static consistency checks. We will formulate alignment as a dynamic arbitration process $\mathcal{O}$ that calculates multi-sensory uncertainty and actively recalibrates weights, enabling a more robust evaluation of how models resolve complex physical and temporal conflicts.

%% file: checklist.tex
\section*{NeurIPS Paper Checklist}

\begin{enumerate}

\item {\bf Claims}
    \item[] Question: Do the main claims made in the abstract and introduction accurately reflect the paper's contributions and scope?
    \item[] Answer: \answerYes{} 
    \item[] Justification: {We discuss the contribution.}

\item {\bf Limitations}
    \item[] Question: Does the paper discuss the limitations of the work performed by the authors?
    \item[] Answer: \answerYes{} 
    \item[] Justification: {We discuss the limitation.}

\item {\bf Theory assumptions and proofs}
    \item[] Question: For each theoretical result, does the paper provide the full set of assumptions and a complete (and correct) proof?
    \item[] Answer: \answerNA{} 
    \item[] Justification: {This paper does not include theoretical results.}

    \item {\bf Experimental result reproducibility}
    \item[] Question: Does the paper fully disclose all the information needed to reproduce the main experimental results of the paper to the extent that it affects the main claims and/or conclusions of the paper (regardless of whether the code and data are provided or not)?
    \item[] Answer: \answerYes{} 
    \item[] Justification: {All experiment results can be reproduced with the instructions in the paper.}

\item {\bf Open access to data and code}
    \item[] Question: Does the paper provide open access to the data and code, with sufficient instructions to faithfully reproduce the main experimental results, as described in supplemental material?
    \item[] Answer: \answerYes{} 
    \item[] Justification: {We release all code, data and details.}

\item {\bf Experimental setting/details}
    \item[] Question: Does the paper specify all the training and test details (e.g., data splits, hyperparameters, how they were chosen, type of optimizer) necessary to understand the results?
    \item[] Answer: \answerYes{} 
    \item[] Justification: {We report all details about the experiments.}

\item {\bf Experiment statistical significance}
    \item[] Question: Does the paper report error bars suitably and correctly defined or other appropriate information about the statistical significance of the experiments?
    \item[] Answer: \answerNo{} 
    \item[] Justification: {All experiments do not need to report error bars.}

\item {\bf Experiments compute resources}
    \item[] Question: For each experiment, does the paper provide sufficient information on the computer resources (type of compute workers, memory, time of execution) needed to reproduce the experiments?
    \item[] Answer: \answerYes{} 
    \item[] Justification: {We report all computing resources.}
    
\item {\bf Code of ethics}
    \item[] Question: Does the research conducted in the paper conform, in every respect, with the NeurIPS Code of Ethics \url{https://neurips.cc/public/EthicsGuidelines}?
    \item[] Answer: \answerYes{} 
    \item[] Justification: {This paper meets the NeurIPS Code of Ethics.}

\item {\bf Broader impacts}
    \item[] Question: Does the paper discuss both potential positive societal impacts and negative societal impacts of the work performed?
    \item[] Answer: \answerNA{} 
    \item[] Justification: {This paper does not contain any impacts on society.}
    
\item {\bf Safeguards}
    \item[] Question: Does the paper describe safeguards that have been put in place for responsible release of data or models that have a high risk for misuse (e.g., pre-trained language models, image generators, or scraped datasets)?
    \item[] Answer: \answerNA{} 
    \item[] Justification: {This paper does not contain any risks.}

\item {\bf Licenses for existing assets}
    \item[] Question: Are the creators or original owners of assets (e.g., code, data, models), used in the paper, properly credited and are the license and terms of use explicitly mentioned and properly respected?
    \item[] Answer: \answerYes{} 
    \item[] Justification: {All resources have been claimed clearly.}

\item {\bf New assets}
    \item[] Question: Are new assets introduced in the paper well documented and is the documentation provided alongside the assets?
    \item[] Answer: \answerYes{} 
    \item[] Justification: {We propose a new dataset and will release it.}

\item {\bf Crowdsourcing and research with human subjects}
    \item[] Question: For crowdsourcing experiments and research with human subjects, does the paper include the full text of instructions given to participants and screenshots, if applicable, as well as details about compensation (if any)? 
    \item[] Answer: \answerNA{} 
    \item[] Justification: {No human subjects.}

\item {\bf Institutional review board (IRB) approvals or equivalent for research with human subjects}
    \item[] Question: Does the paper describe potential risks incurred by study participants, whether such risks were disclosed to the subjects, and whether Institutional Review Board (IRB) approvals (or an equivalent approval/review based on the requirements of your country or institution) were obtained?
    \item[] Answer:  \answerNA{} 
    \item[] Justification: {No human experiment.}

\item {\bf Declaration of LLM usage}
    \item[] Question: Does the paper describe the usage of LLMs if it is an important, original, or non-standard component of the core methods in this research? Note that if the LLM is used only for writing, editing, or formatting purposes and does \emph{not} impact the core methodology, scientific rigor, or originality of the research, declaration is not required.
    \item[] Answer: \answerYes{} 
    \item[] Justification: {We benchmark some LLMs and report their performance.}

\end{enumerate}